\documentclass[10pt,conference]{IEEEtran}
\IEEEoverridecommandlockouts
\usepackage{cite}
\usepackage{amsmath,amssymb,amsfonts}
\usepackage{algorithmic}
\usepackage{graphicx}
\usepackage{textcomp}
\usepackage{xcolor}
\usepackage{multirow} 
\def\BibTeX{{\rm B\kern-.05em{\sc i\kern-.025em b}\kern-.08em
    T\kern-.1667em\lower.7ex\hbox{E}\kern-.125emX}}
\begin{document}

\title{Causal Debiasing for Visual Commonsense Reasoning%Conference Paper Title*\\

\thanks{
*corresponding author.

This work was supported by the National Natural Science Foundation of China under Grants (No.62325206, 62306150), and the Key Research and Development Program of Jiangsu Province under Grant BE2023016-4.}
}

\author{\IEEEauthorblockN{Jiayi Zou}
\IEEEauthorblockA{
\textit{Nanjing University of}\\
\textit{Posts and Telecommunications} \\
Nanjing, China \\
2023010213@njupt.edu.cn}
\and
\IEEEauthorblockN{Gengyun Jia}
\IEEEauthorblockA{
\textit{Nanjing University of}\\
\textit{Posts and Telecommunications} \\
Nanjing, China \\
gengyun.jia@njupt.edu.cn}
\and
\IEEEauthorblockN{Bing-Kun Bao*}
\IEEEauthorblockA{
\textit{Nanjing University of}\\
\textit{Posts and Telecommunications} \\
Nanjing, China \\
bingkunbao@njupt.edu.cn}

}

\maketitle
\textbf{}
\begin{abstract}
Visual Commonsense Reasoning (VCR) refers to answering questions and providing explanations based on images. While existing methods achieve high prediction accuracy, they often overlook bias in datasets and lack debiasing strategies. In this paper, our analysis reveals co-occurrence and statistical biases in both textual and visual data. We introduce the VCR-OOD datasets, comprising VCR-OOD-QA and VCR-OOD-VA subsets, which are designed to evaluate the generalization capabilities of models across two modalities. Furthermore, we analyze the causal graphs and prediction shortcuts in VCR and adopt a backdoor adjustment method to remove bias. Specifically, we create a dictionary based on the set of correct answers to eliminate prediction shortcuts. Experiments demonstrate the effectiveness of our debiasing method across different datasets.
\end{abstract}


\section{Introduction}
The Visual Commonsense Reasoning (VCR) \cite{zellers2019recognition} task is an extension of Visual Question Answering (VQA) \cite{feng2024rankdvqa, patel2021generating} and consists of two sub-tasks: four-choice question answering (Q$\rightarrow$A) and reasoning (QA$\rightarrow$R). In the Q$\rightarrow$A sub-task, the model selects the correct answer from four options based on the image. In the QA$\rightarrow$R sub-task, the model selects the appropriate reasoning for the chosen answer. 

Existing research, including task-specific \cite{yu2019heterogeneous, zhang2021explicit, zhang2021multi, li2023joint, zhu2023multi} and pre-training models \cite{chen2020uniter, wang2022sgeitl, cho2021unifying, yuan2023gpt, ma2024eventlens, cai2024vip}, has made a lot of progress in VCR.
Task-specific models use attention mechanisms \cite{song2020deep, lu2023multi, sood2023multimodal} or graph networks \cite{song2023efficient, wang2023vqa, zhou2024graph} to capture image and question correlations for accurate answering and reasoning. Pre-trained models leverage prior training on other datasets and fine-tuning on VCR, benefiting from learned relationships and enhancing visual-textual alignment \cite{khan2022common, cheng2022gsrformer}. 

Despite significant advancements in VCR, researchers have found that models may rely on biases rather than reasoning capabilities to answer questions \cite{li2023joint, vosoughi2024cross}. The issue is initially investigated by Ye et al. \cite{ye2021case}, who find that models heavily depend on co-occurrence for predictions when encountering repeated references. However, their study focus solely on the co-occurrence of person tags and do not extensively analyze other types of bias. Subsequent research has not adequately addressed the challenge of bias mitigation \cite{kim2023dynamic}. 

To bridge this gap, our study investigates bias challenges in VCR, focusing on co-occurrence bias and statistical bias. Co-occurrence bias arises from co-occurring components between modalities. During multi-modal feature fusion, the model may pay more attention to the common words between questions and answers or the co-occurrence between visual objects and answers, leading to biased predictions. To validate this, we conduct experiments using the R2C \cite{zellers2019recognition} on segmented datasets with or without co-occurring words between questions and answers. The results in Table I confirm that the baseline model achieves higher accuracy on the co-occurrence split, supporting our hypothesis.

Moreover, we observe that certain question types and image types can introduce statistical bias. For example, when questions include specific verbs, the frequency of words in the ground-truth answers may vary. In such cases, the model may rely excessively on high-frequency words and undermine visual information. To investigate this, we divide the dataset based on the frequency of words in answers for each question type. This results in two subsets: one with high-frequency words (head-labeled) and another with low-frequency (tail-labeled). Experiments in Table I show that accuracy of Q$\rightarrow$A  is higher on the head-labeled dataset, providing evidence for statistical bias.

\begin{table}[]
\caption{The R2C model predicts accuracy(\%) on different split data sets to verify the existence of two biases.
} \label{tab:cap}
\begin{center}
%\resizebox{\linewidth}{!}{
\begin{tabular}[width=1.0\linewidth]{|l|ll|}
\hline
Subsets& Q$\rightarrow$A&   QA$\rightarrow$R\\ \hline
co-occurrence&      36.9&     44.4\\
 None co-occurrence& 31.3& 30.9\\ \hline
 head-labeled& 29.6& --\\
                                                                       tail-labeled&      27.9&     --\\ \hline
\end{tabular}%}
\end{center}
\end{table}

After analyzing the bias, we introduce the out-of-distribution (OOD) VCR-OOD dataset, which deals with both co-occurrence bias and statistical bias. For the text modality, we create VCR-OOD-QA by filtering samples with co-occurring words between questions and answers, and balancing verb frequency in answers for different question types. For the visual modality, we apply similar procedures to obtain VCR-OOD-VA. 

To combat bias, we develop a debiasing model based on causal intervention \cite{li2022causal, wang2020visual}. Given the causal graph, we depict the causal relationships in VCR and analyze the prediction shortcuts provided by confounding factors that lead to model learning bias. By employing the backdoor adjustment method, we design an answer dictionary to disrupt the propagation path of these confounding factors and eliminate bias. The backdoor adjustment approach allows us to adjust for the effect of confounding factors by conditioning on variables that block the spurious associations between the input and the output. We create the dictionary based on the correct answer set, ensuring that the model focuses on the relevant information rather than relying on shortcuts that might arise due to biases in the training data.

In summary, our contributions can be categorized into three main areas: 
\begin{itemize}
\item We analyze two types of biases in VCR, including co-occurrence bias and statistical bias, and design VCR-OOD datasets. 
\item In VCR, we utilize causal graphs to capture the influence of confounding factors on predictions. 
\item To cut off the shortcut and alleviate bias, we introduce a debiasing method based on the backdoor adjustment to correct multi-modal features. 
\end{itemize}

\section{Construction of OOD datasets}
 
In this section, we leverage the in-distribution (ID) dataset VCR \cite{zellers2019recognition} to create the VCR-OOD dataset, which comprises two subsets: VCR-OOD-QA for the text modality and VCR-OOD-VA for the visual modality. Below we will introduce our processing method in detail.
%\subsection{VCR-OOD-QA}

For the text modality, the VCR-OOD-QA dataset is constructed with the focus on co-occurrence bias and statistical bias. We filter out samples where there are co-occurring words between the questions and the correct answers, as well as between the connected question-answer pairs and the correct reasoning. This process create an auxiliary dataset. To tackle statistical bias, we first categorize the questions based on the verbs they contain, forming different question sets ($Q_{1}$, $Q_{2}$, etc.). We then record the corresponding correct answer sets ($A_{1}$, $A_{2}$, etc.) and extract the verb or noun from them. Based on the frequency of these words, we divide the samples for each question type into "head-labeled" (high-frequency) and "tail-labeled" (low-frequency) subsets. The division threshold, denoted as $\alpha$, corresponds to the frequency of the word ranked in the middle. Finally, we randomly retain a threshold number of "head-labeled" samples from each question type to create the VCR-OOD-QA dataset. To maintain dataset consistency, we randomly select 10,000 training samples to balance the 1,045 validation samples. 

In the visual modality, establishing correspondence between objects and answer text can be challenging. However, as many questions describe objects present in the images, we leverage the auxiliary dataset in VCR-OOD-QA to mitigate this bias. We generate VCR-OOD-VA using a similar methodology, but categorize images based on the contained objects rather than question text. This results in image sets $I_{1}$, $I_{2}$, etc. We then apply the head-tail division and sampling approach used for VCR-OOD-QA to construct the 1,687-sample VCR-OOD-VA validation set, while the training set remains the same as VCR-OOD-QA. 

\section{Our Method}
To address the bias issue, we analyze the causal relationships \cite{zhang2023reducing} and present the shortcuts in the causal graph\cite{zheng2022causally}. Then, we adopt the backdoor adjustment method \cite{zhang2023backdoor} to debias. 
%kim2023survey,
\begin{figure}
    \centering
    \includegraphics[width=1.0\linewidth]{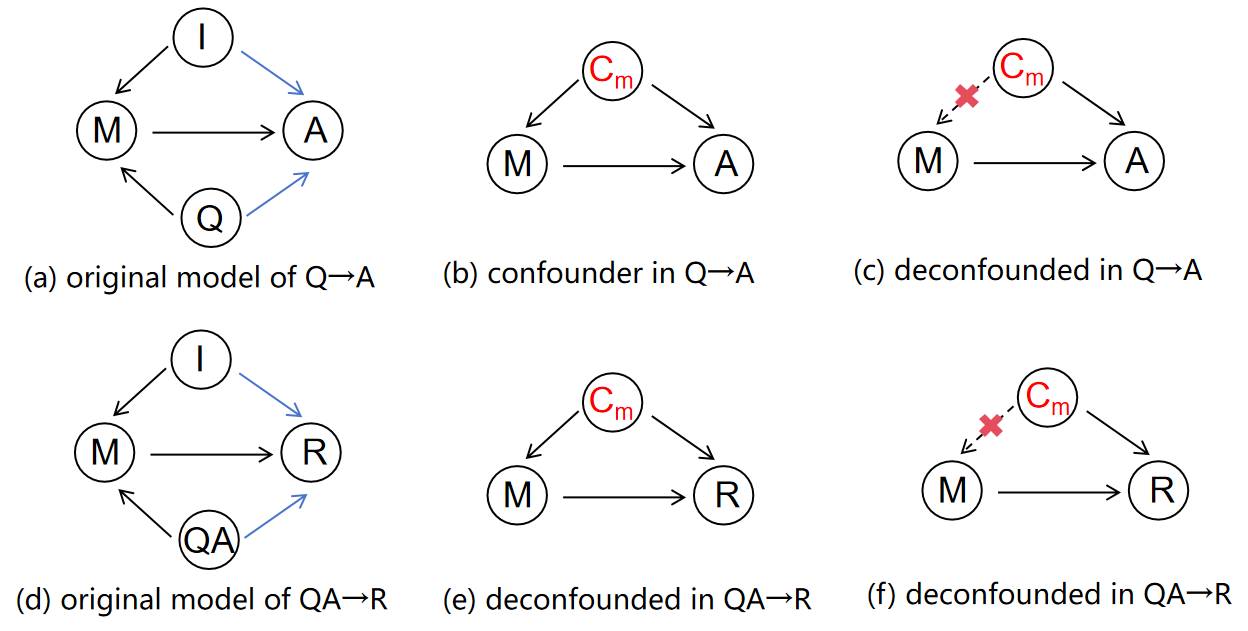}
    \caption{(a) is the causal graph in Q$\rightarrow$A . (b) shows the confounder in Q$\rightarrow$A. (c) is our deconfounded in Q$\rightarrow$A. (d), (e) and (f) correspond to the representations in QA$\rightarrow$R respectively. The blue arrows indicate the prediction shortcuts.
}
    \label{fig:enter-label}
\end{figure}
%qa到r
%因为有不正常的路径，用一个字典切断两条路径
Causal graphs are directed acyclic graphs (DAGs) \cite{qi2020two} consisting of vertices (V) and edges (E), representing causal relationships between variables. Figure 1 (a) and (d) show the causal graphs of two sub-tasks in VCR, including image (I), question (Q), fused features (M), predicted answer (A), connecting question and true answer (QA), and predicted reasoning (R). Among them, the blue arrow indicates the existence of prediction shortcuts, such as I$\rightarrow$A. In (b) and (e), we introduce a hybrid variable $C_{m}$ to represent the effects of co-occurrence and statistical regularity. The direct causal effect from M to A captures the influence of the image-text features on the predicted answers. However, the causal path $M\leftarrow C_{m}\rightarrow A$ indicates that confounding factor creates a backdoor path originating from M, leading to a spurious correlation between M and A. 
The predicted probability can be expressed as $P(A|M)=\sum_{c\in C_{m}}P_{c}(A|M,c)P(c|M)$. 

As shown in Figure 1 (c) and (f), we propose a causal intervention method, do-calculus \cite{pearl2018book}, to cut off the edge between $C_{m}$ and $M$. 
%We will use the Q$\rightarrow$A task as an example to introduce the theory of causal analysis. 
By considering all possible values of the confounding factor, we can estimate the predicted probability of a candidate answer using a backdoor adjustment approach:
\begin{equation}
\begin{split}
P(A|do(M))& = \sum_{c\in C_{m}}P_{c}(A|M,c)P(c)\\
%& = \mathbb{E}_{c\in C_{m}}[P_{c}(A|M,c)]
\end{split}
\end{equation}
In typical models, the last layer often consists of a linear softmax structure, where the predicted probability given the multi-modal features $m$ processed through the encoder $f()$ is computed as $P_{c}(A|M) = \text{softmax}(f(m))$. 
To incorporate the confounding factor into this process, we directly include it in the feature processing during the softmax calculation, resulting in $P_{c}(A|M,c) = \text{softmax}(f(m),c)$. 
Then, we employ the Normalized Weighted Geometric Mean (NWGM) approximation \cite{krejvci2018aggregation} to simplify the representation of $P(A|do(M))$. Taking into account the independence between the multi-modal features $m$ and $c\in C_{m}$, Formula(1) is simplified as:
\begin{equation}
\begin{split}
& P(A|do(M))\\
%& \mathbb{E}_{c\in C_{m}}[P_{c}(A|M,c)]\\
=& \mathbb{E}_{c\in C_{m}}[\text{softmax}(f(m),c)]\\
\approx&\text{softmax}[ \mathbb{E}_{c\in C_{m}}(f(m)+c)]\\
=&\text{softmax}[f(m)+ \mathbb{E}_{c\in C_{m}}(c)]
\end{split}
\end{equation}
where $\mathbb{E}_{c\in C_{m}}$ is the mathematical expectation with respect to $c\in C_{m}$.

Figure 2 illustrates the framework of our model, with a primary focus on approximating confounding factors. To achieve this, we design a dictionary $D[C_{m}]$ of dimensions $N\times d$, where $N$ denotes the predetermined number of candidate answers, and $d$ represents the hidden layer size of the model.  Specifically, we randomly select $N$ correct answers from the training set and feed them into the text encoder, integrating them into the dictionary. We construct corresponding dictionaries for the two sub-tasks of VCR. Thanks to the linear additive property of expectation computation, we can use $f(m)+ \mathbb{E}_{c\in C_{m}}(c)$ to compute $ \mathbb{E}_{c\in C_{m}}(f(m)+c)$. Specifically, $\mathbb{E}_{c\in C_{m}}(c)=softmax (L^\mathsf{T}K)D[C_{m}]$, where $L = W_{1}f(m), K = W_{2}D[C_{m}]$ are the product of elements, and $W_{1}$, $W_{2}$ are the mapping matrices. 

\begin{figure}
    \centering
    \includegraphics[width=1.0\linewidth]{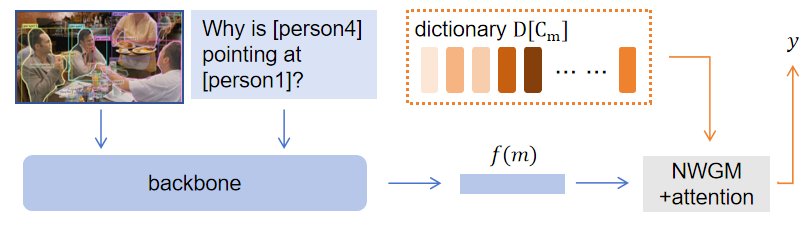}
    \caption{The framework diagram of causal debiasing in the model. The blue arrows represent the prediction process of the backbone, and the orange arrows are the operations that introduce the do operator.}
    \label{fig:enter-label}
\end{figure}

Considering that the answer dictionary is a feature of the text modality, the model may ignore the information of the visual modality. In order to further enhance the model's ability to focus on visual content, we introduce negative samples. For each sample triple $(Q, I, A)$ in Q$\rightarrow$A, including query, image and answer, we randomly select an image in the batch to form a negative sample $(Q, I^-, A)$, and input it into the model to calculate the loss. 
\begin{eqnarray}
Loss_{neg}=-\sum_{i=1}^4y_{i}log(f_{base}(Q,I^{-},A))
\end{eqnarray}
where $i$ is the sample index, $y$ is the ground truth, and $f_{base}$ indicate the predicted value obtained from the baseline model. For each sample triple $(QA, I, R)$ in Q$\rightarrow$A, Formula 3 can be rewritten as:
\begin{eqnarray}
Loss_{neg}=-\sum_{i=1}^4y_{i}log(f_{base}(QA,I^{-},R))
\end{eqnarray}
The main distinction between this negative loss function and the baseline loss $loss_{base}$ \cite{li2023learning} lies in the input image.   This loss serves as a penalty to incentivize the model to concentrate on the accurate visual content. Hence, the training loss can be formulated as follows:
\begin{eqnarray}
Loss=loss_{base}+\lambda*loss_{neg}
\end{eqnarray}
where $\lambda$ is a hyper-parameter.
\section{Experiments}
\begin{table}[]
\caption{Accuracy (\%) of baseline methods and our method on the VCR-OOD-QA validation set. The highest accuracy results are displayed in bold and black, and the second-best results are underlined.
} \label{tab:cap}
\begin{center}
\begin{tabular}{c|c|c|c}
\hline
Methods& Q$\rightarrow $A&QA$\rightarrow$R& Q$\rightarrow $AR\\ \hline
 R2C \cite{zellers2019recognition}& 30.3& 31.0& 9.7\\
 HGL \cite{yu2019heterogeneous}& 36.3& 36.7& 14.7\\
 MCC \cite{zhang2021multi}& \underline{40.4}& 34.4&14.0\\
 ARC \cite{li2023joint}& 33.6& 37.8&14.7\\
MSGT \cite{zhu2023multi}& 37.7& \underline{38.1}& \underline{15.7}\\ \hline
 R2C+ours& 34.4& 37.2&13.7\\
 MSGT+ours& \textbf{42.2}& \textbf{40.3}&\textbf{18.4}\\\hline
\end{tabular}
\end{center}
\end{table}

\begin{table}[]
\caption{Accuracy (\%) of baseline methods and our method on the VCR-OOD-VA validation set. The highest accuracy results are displayed in bold and black, and the second-best results are underlined.
} \label{tab:cap}
\begin{center}
\begin{tabular}{c|c|c|c}
\hline
Methods& Q$\rightarrow $A&QA$\rightarrow$R& Q$\rightarrow $AR\\ \hline
 R2C \cite{zellers2019recognition}& 32.2& 34.0& 10.8\\
 HGL \cite{yu2019heterogeneous}& 41.8& 38.1& 17.4\\
 MCC \cite{zhang2021multi}& \underline{46.6}& \underline{38.9}&\underline{18.5}\\
 ARC \cite{li2023joint}& 37.8& 37.5&15.5\\
MSGT \cite{zhu2023multi}& 46.5& 35.1& 16.6\\ \hline
 R2C+ours& 34.8& 38.4&15.0\\
 MSGT+ours& \textbf{47.3}& \textbf{40.3}&\textbf{19.2}\\\hline
\end{tabular}
\end{center}
\end{table}
\subsection{Experimental Setup}
We conduct extensive experiments on VCR \cite{zellers2019recognition} and our VCR-OOD dataset, using three commonly used evaluation metrics Q$\rightarrow$A, QA$\rightarrow$R and Q$\rightarrow$AR.  The ID dataset VCR contains 290,000 four-option multiple-choice questions from 110,000 movie scenes. In the experiment, we select some VCR task-specific models as competitors, such as R2C \cite{zellers2019recognition}, HGL \cite{yu2019heterogeneous}, MCC \cite{zhang2021multi}, ARC \cite{li2023joint}, and MSGT \cite{zhu2023multi}. 
We train and test the above 5 classic VCR models on VCR-OOD. %Among them, R2C \cite{zellers2019recognition} is the earliest model proposed in VCR, using LSTM structure for hierarchical reasoning. MSGT \cite{zhu2023multi} introduces a graph network structure to integrate multi-modal features. 
Since our method is model-independent, we add our method to R2C \cite{zellers2019recognition} and MSGT \cite{zhu2023multi} for comparison. To further verify the effectiveness of each module of the model, we conduct ablation experiments. The spaCy toolkit \cite{honnibal2017spacy} was used to extract verbs and nouns from the text in the VCR-OOD dataset. The answer features in the dictionary are obtained by Bert. We artificially set the size of the dictionary, where $N$ is 1000 and $d$ is 512. For a fair comparison, our parameter settings follow those reported in the original article.

\subsection{Quantitative Result}
In this section, we evaluate five task-specific models on VCR-OOD and then compare our method's performance against the baseline models R2C \cite{zellers2019recognition} and MSGT \cite{zhu2023multi}. 
In addition, we also conduct experiments on the ID dataset VCR \cite{zellers2019recognition}.

\begin{table}[]
\caption{Accuracy (\%) of our method based on MSGT on the VCR validation set, which is an ID not OOD dataset.
} \label{tab:cap}
\begin{center}
\begin{tabular}{c|c|c|c}
\hline
Methods& Q$\rightarrow $A&QA$\rightarrow$R& Q$\rightarrow $AR\\ \hline
 R2C \cite{zellers2019recognition}& 63.8& 67.2& 43.1\\
 HGL \cite{yu2019heterogeneous}& 69.4& 70.6& 49.1\\
 MCC \cite{zhang2021multi}& 71.7& 73.4&52.9\\
 ARC \cite{li2023joint}& 70.5& 73.1&51.8\\
MSGT \cite{zhu2023multi}& 72.2& 73.6& 53.3\\\hline
 MSGT+ours& 71.3& 72.9&52.2\\\hline
\end{tabular}
\end{center}
\end{table}

Table II shows the results of the 5 baseline models and our method on VCR-OOD-QA. We find that the graph-based models HGL \cite{yu2019heterogeneous} and MSGT \cite{zhu2023multi} outperform the basic R2C \cite{zellers2019recognition} model. While the contrastive learning-based MCC \cite{zhang2021multi} achieves the highest accuracy of 40.4\% on Q$\rightarrow$A, it struggles on reasoning tasks, scoring only 14.0\% on Q$\rightarrow$AR. ARC matches HGL's \cite{yu2019heterogeneous} 14.7\% on Q$\rightarrow$AR. Notably, MSGT \cite{zhu2023multi} achieves the highest accuracies among the baselines, with 38.1\% on QA$\rightarrow$R and 15.7\% on Q$\rightarrow$AR. 
We implement our method on R2C \cite{zellers2019recognition} and MSGT \cite{zhu2023multi} respectively. On R2C \cite{zellers2019recognition}, after applying our method, the prediction results are improved compared with the baseline in three indicators. After using MSGT \cite{zhu2023multi} as the baseline, our method achieve the best in all three indicators, with prediction accuracy of 42.2\%, 40.3\% and 18.4\%.

The results on VCR-OOD-VA are shown in Table III. HGL \cite{yu2019heterogeneous} shows significant improvement over R2C \cite{zellers2019recognition} across all three metrics. MCC \cite{zhang2021multi} achieves the highest accuracy, with 46.6\% on Q$\rightarrow$A, 38.9\% on QA$\rightarrow$R, and 19.0\% on Q$\rightarrow$AR, exceeding R2C \cite{zellers2019recognition} by 9.8 points on the latter. While the distillation in ARC does not work well on this dataset, scoring only 15.5\% on Q$\rightarrow$AR, MSGT \cite{zhu2023multi} achieves 46.5\% on Q$\rightarrow$A. Importantly, our method outperforms MSGT \cite{zhu2023multi}, reaching the highest accuracies of 47.3\% on Q$\rightarrow$A, 40.3\% on QA$\rightarrow$R, and 19.2\% on Q$\rightarrow$AR. 

On the ID dataset VCR, after adding our debiasing method, the prediction accuracy of the model is slightly lower than the baseline, but it is still competitive. Table IV shows that our method achieves a prediction accuracy of 52.2\% on Q$\rightarrow$AR. 
Our debiasing method does not completely damage the performance of the model, which is consistent with other debiasing works  \cite{zhang2023next, lu2024causalme,lv2024modality} in VQA. 
%We also observe that the accuracy on our proposed VCR-OOD dataset is generally lower compared to that on VCR. This may be because our OOD dataset retains samples that are less common and thus more challenging for the model to handle accurately.

\subsection{Ablation Study}
To evaluate the effectiveness of each component and the influence of $\lambda$ in the loss, we utilize MSGT \cite{zhu2023multi} as the baseline and perform ablation experiments on VCR-OOD. %Additionally, we investigate the influence of the hyper-parameter $\lambda$ in the loss. 

The results of each component on VCR-OOD-QA and VCR-OOD-VA are presented in Table V. Compared to the baseline, the model demonstrates improved performance on each sub-task when incorporating causal intervention. 
%This indicates the effectiveness of the backdoor adjustment debiasing method. 
When both components are combined, the model achieves the best performance with accuracy rates of 18.4\% and 19.2\% on VCR-OOD-QA and VCR-OOD-VA, respectively. This suggests that both parts further enhance the model's prediction accuracy. %$loss_{neg}$ compensates for the limitations of causal intervention and

\begin{table}[]
\caption{Ablation study on the validation set of VCR-OOD-QA and VCR-OOD-VA dataset using MSGT as the baseline. 
} \label{tab:cap}
%\begin{center}
\resizebox{\linewidth}{!}{
\begin{tabular}{lcc|ccc}
\hline
   dataset&causal inference
&$loss_{neg}$& Q$\rightarrow $A&QA$\rightarrow$R& Q$\rightarrow $AR\\ \hline
    &$\times$&$\times$& 37.7& 38.1& 15.7\\
    \multirow{2}*{VCR-OOD-QA}&\checkmark&$\times$& 41.2& 38.3& 16.4\\
    ~&$\times$&\checkmark& 41.9& 36.3&16.0\\
   &\checkmark&\checkmark& 42.2& 40.3& 18.4\\\hline
  &$\times$& $\times$& 46.5& 35.1&16.6
\\
   \multirow{2}*{VCR-OOD-VA}&\checkmark& $\times$& 47.0& 35.6&16.8
\\
  ~&$\times$& \checkmark& 46.3& 38.6&17.9
\\
  &\checkmark& \checkmark& 47.3& 40.3&19.2\\\hline
\end{tabular}}
%\end{center}
\end{table}

\begin{figure}
    \centering
    \includegraphics[width=1.0\linewidth]{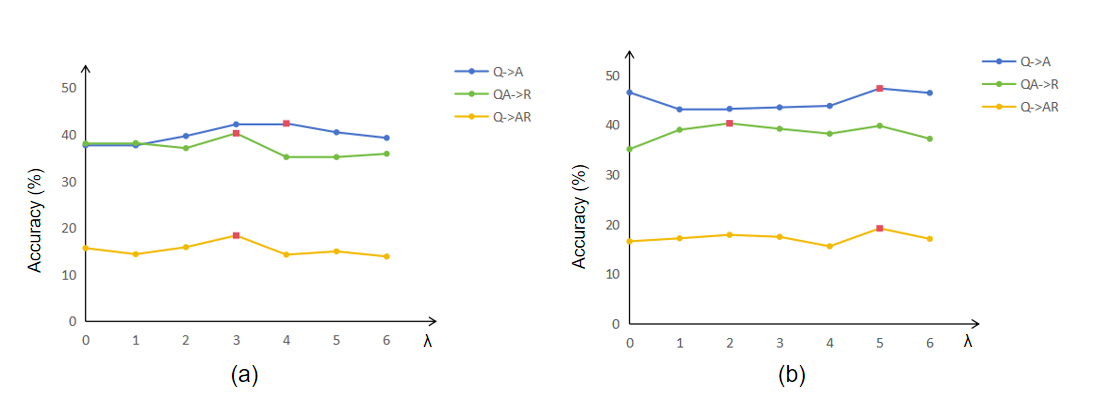}
    \caption{Performance changes with different parameter values ($\lambda$) on VCR-OOD-QA (a) and VCR-OOD-VA Datasets (b).}
    \label{fig:enter-label}
\end{figure}

Furthermore, we investigate the impact of the hyper-parameter $\lambda$ in the loss function as shown in Figure 3. The accuracy of Q→AR is evaluated by correctly predicting the prediction results on both Q→A and Q→AR with the same $\lambda$ value. On VCR-OOD-QA, the model achieves the highest accuracy of 40.3\% on QA$\rightarrow$R when $\lambda$ is 4, and 42.2\% on Q$\rightarrow$A, 18.4\% on Q$\rightarrow$AR when $\lambda$ is 3.  On the VCR-OOD-VA dataset, the accuracy reaches the highest in terms of Q$\rightarrow$A, QA$\rightarrow$R and Q$\rightarrow$AR when $\lambda$ values are 5, 2, 5 respectively. 

\section{Conclusion}
In this paper, we conduct a comprehensive analysis of the bias issue in the VCR task, considering both text and vision modalities. To evaluate the model's generalization, we propose the VCR-OOD dataset. To address the issue of bias learning, we begin by constructing a task-specific causal graph that incorporates bias effects. By employing the backdoor adjustment method in causal intervention, we design an answer dictionary to disrupt the prediction shortcuts. Finally, we evaluate the performance of our proposed method on both ID and OOD datasets. 

\vfill\pagebreak


\bibliographystyle{IEEEbib}
\bibliography{refs}

\end{document}